\documentclass[10pt,twocolumn,letterpaper]{article}

\usepackage{cvpr}
\usepackage{times}
\usepackage{epsfig}
\usepackage{graphicx}
\usepackage{amsmath}
\usepackage{amssymb}
\usepackage[normalem]{ulem}
\usepackage{afterpage}
\usepackage{booktabs}
\usepackage{color}
\usepackage{verbatim}
\usepackage{soul}
\usepackage{xspace}
\usepackage{comment}
\usepackage{array}
\usepackage{multirow}
\usepackage{url}
\usepackage{epigraph}
\usepackage[toc,page]{appendix}
\usepackage{microtype}

\cvprfinalcopy

\usepackage[pagebackref=true,breaklinks=true,letterpaper=true,colorlinks,bookmarks=false]{hyperref}

\makeatletter
\DeclareRobustCommand\onedot{\futurelet\@let@token\@onedot}
\def\@onedot{\ifx\@let@token.\else.\null\fi\xspace}

\def\etal{\emph{et al}\onedot}

\makeatother
\newcommand{\thickhline}{%
    \noalign {\ifnum 0=`}\fi \hrule height 1pt
    \futurelet \reserved@a \@xhline
}

\newcolumntype{"}{@{\hskip\tabcolsep\vrule width 1.5pt\hskip\tabcolsep}}

\usepackage{booktabs}

\ifcvprfinal\fi

\title{Improving task-specific representation via 1M unlabelled images \\without any extra knowledge} 
\author{Aayush Bansal\\
Carnegie Mellon University\\
{\tt\small{aayushb@cs.cmu.edu}}
}

\begin{document}

\maketitle

\begin{abstract}

We present a case-study to improve the task-specific representation by leveraging a million unlabelled images without any extra knowledge. We propose an exceedingly simple method of conditioning an existing representation on a diverse data distribution and observe that a model trained on diverse examples acts as a better initialization. We extensively study our findings for the task of surface normal estimation and semantic segmentation from a single image. We improve surface normal estimation on NYU-v2 depth dataset and semantic segmentation on PASCAL VOC by $4$\% over base model. We did not use any task-specific knowledge or auxiliary tasks, neither changed hyper-parameters nor made any modification in the underlying neural network architecture. 

\end{abstract}


\section{Introduction}
\label{sec:intro}

We present a simple approach to improve a task-specific representation by using a million unlabelled images without any extra knowledge. Better task-specific representation leads to improved performance on the task-of-interest. As such, each one of us aspire for a better performing models for a task we care about. There are three standard ways to learn a better representation: (1) using more labelled data~\cite{MSCOCO-2014,Russakovsky15,zhou2017places}; (2) using better convolutional neural network (CNN) architectures~\cite{he2015deep,huang2017densely,krizhevsky2012imagenet,SimonyanZ14a} or finding better architectures~\cite{cao2018learnable,tan2019efficientnet,zoph2018learning}; (3) adding task-specific domain knowledge~\cite{qi2018geonet,Wang15} or using auxiliary tasks~\cite{doersch2015unsupervised,Gidaris2018,WangG15,rzhang2016colorful}. 

\begin{figure}[t]
\includegraphics[width=\linewidth]{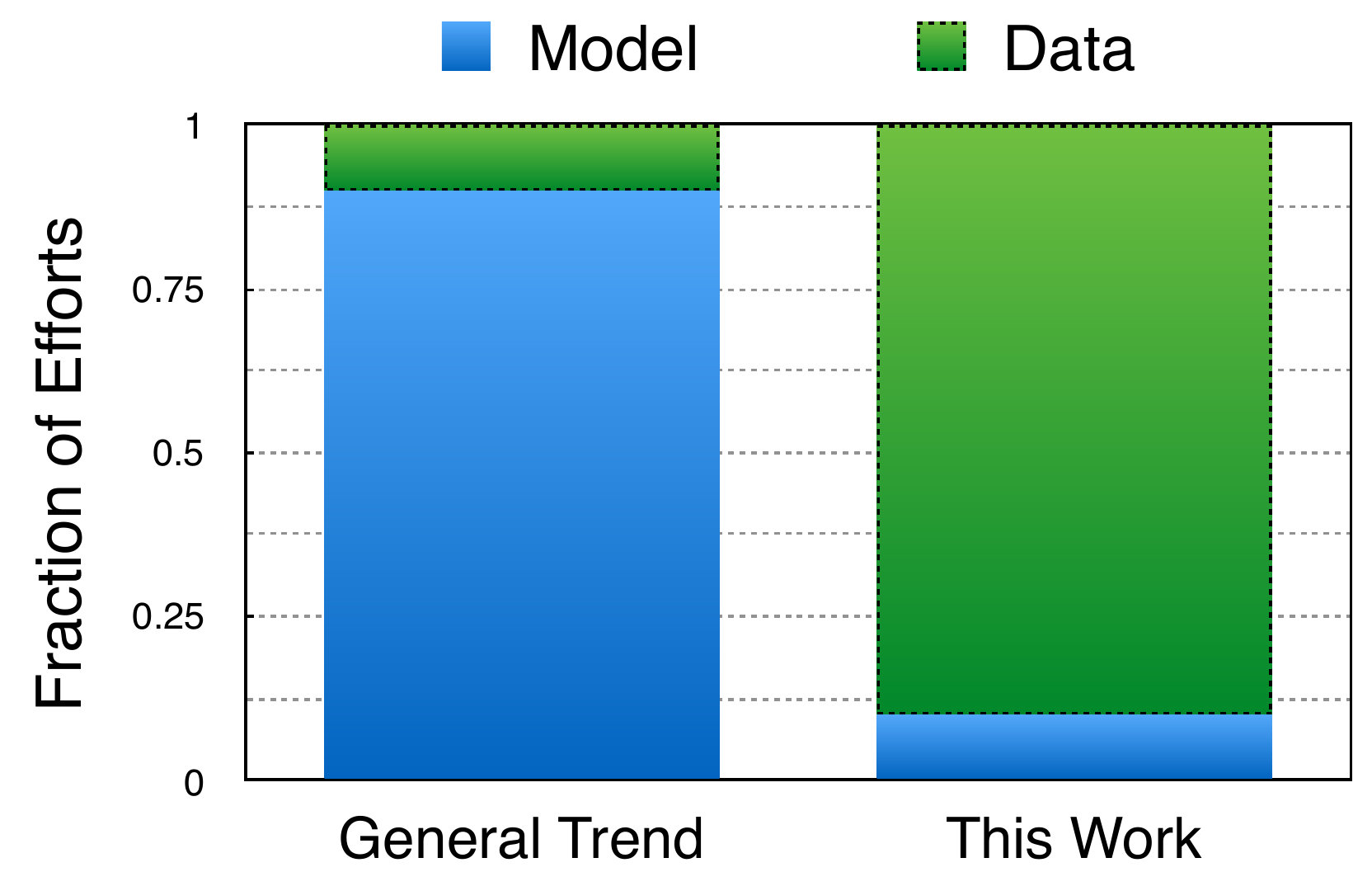}
\caption{\textbf{General Trend vs. This Work: } Recent work in computer vision literature primarily focuses on designing better CNN-architectures and optimization to improve the performance of tasks on various benchmarks. In this work, our focus is to use the \emph{freely-available} unlabelled images in the simplest possible way to learn a better representation. Only effort spend on the model part is to look-up the hyper-parameters from prior work and use them to train a new model with a million unlabelled images.}
\label{fig:teaser}
\end{figure}

\textbf{Our Work: } We take a less explored fourth way to improve a  task-specific representation, trained from small datasets, by naively using a million unlabelled images. We do this by conditioning an existing representation on a diverse data distribution. Suppose, we are given two completely different data distribution in our setup: NYU-v2 depth dataset~\cite{Silberman12}  and  ImageNet~\cite{Russakovsky15}.  NYU-v2 depth dataset has an image and surface normal map pair : $\{(X_1,Y_1)\}$. We do not have any other information for a million images from ImageNet ($X_{2}$). Firstly, we learn a mapping $F:X_{1}{\rightarrow}Y_{1}$. We use the mapping $F$ to predict surface normal map for $X_2$, and learn a new mapping $G:X_{2}{\rightarrow}F(X_{2})$.  We demonstrate that $G$, when fine-tuned for $\{(X_{1}, Y_{1})\}$, leads to better performance than $F$ that is trained using exactly same parameters and computational resources. This observation implies that a mapping learnt using a diverse data distribution can act as a better initialization even when we do not have any labels or use additional information.

We use the task of surface normal estimation for our demonstration because NYU-v2 depth dataset and ImageNet are completely different data distribution. It is counter intuitive~\cite{CVPR11_Torralba} as how the predicted surface normal map for a million images from ImageNet can help in improving the performance of surface normal estimation on NYU-v2 depth dataset. We also extend our study for semantic segmentation and observe similar behavior. In this work, we primarily study the role of unlabelled visual data in learning a better representation. As shown in Figure~\ref{fig:teaser}, we spend a major fraction of our efforts in analyzing the influence of data and do not make any innovation on optimization or network architecture.

\textbf{Auxiliary Tasks: } The different approaches for learning a representation in self-supervised manner~\cite{doersch2015unsupervised,Gidaris2018,rzhang2016colorful} define an auxiliary task. While most of these approaches are using the unlabelled images from ImageNet, it is not clear if the performance improvement is due to the task or the images. One may argue as how can we get a better performance than what we already have without any extra supervision or an auxiliary source of supervision? We demonstrate that we can indeed learn a better representation for a task without any induced knowledge by leveraging a million unlabelled images. We improve surface normal estimation on NYU-v2 depth dataset and semantic segmentation on PASCAL VOC by $4$\% over the base model. Importantly, we propose a careful study that aims to isolate the \emph{influence of visual data} amongst other factors involved in learning a representation. 

\textbf{Weakly or Semi Supervised Learning:} The power of data has also been explored in weakly supervised learning~\cite{Izadinia:2015,joulin2016learning,sun2017revisiting} where weak labels (such as user tags etc) are provided, or in a semi-supervised setting~\cite{MisraSSL15,Radosavovic2017,zhang2016augmenting} with a few labeled data and largely unlabelled data. Our work is partially inspired from these weakly-supervised and semi-supervised approaches as we try to simulate labels on a diverse set of unlabelled images to learn a better representation. Different from weakly supervised approaches, we do not use any additional source of knowledge with the images. Finally, our work shares similarity with recently proposed data distillation approach by Radosavovic et al.~\cite{Radosavovic2017}. Our approach is exceedingly simple. We do not assume a good teacher model in our work. The initial model that is used to simulate labels is trained using a small dataset. Despite this, we see a similar performance improvement for the tasks of surface normal estimation and semantic segmentation.

\section{Method}
\label{sec:approach}

\begin{figure*}[t]
\includegraphics[width=\linewidth]{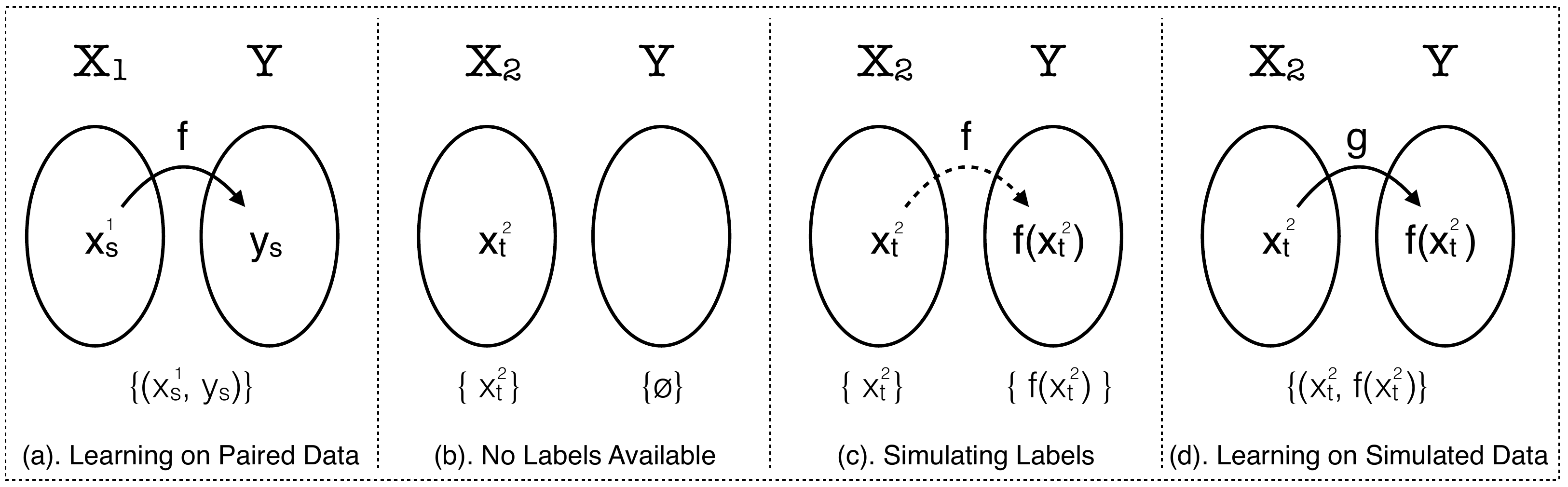}
\caption{\textbf{Conditioning representation on a different data distribution: }  The figure qualitatively shows how we condition a learnt representation on a different data distribution $X_2$ that has no labeled data.  As shown in \textbf{(a)}, we learn a mapping function $f$ from a paired data $\{(x_{s}^{1},y_{s})\}$. Since there exists no labeled data for $X_2$ (as shown in \textbf{(b)}), we generate labels via $f$ (as shown in \textbf{(c)}). Finally \textbf{(d)}, we use this image and simulated label data pair to learn another visual representation Shown in Eq.~\ref{eq:reg2}, $g$ is trying to \emph{mimic} $f$ via samples in $X_2$.}
\label{fig:overview}
\end{figure*}

A fundamental goal of this work is to isolate the influence of visual data, $X$, for learning a mapping $f:X \rightarrow Y$ where $Y$ is the intended target. The samples in X are images ($H{\times}W{\times}3$) and samples in Y is the target task ($H{\times}W{\times}N$). $H$ and $W$ are the height and width of an image, and $N$ is the dimension for the  target.

\noindent\textbf{Data: } We poke $X$ by varying its source and distribution. $X_1$ and $X_2$ are two data sources, and each comes from a different distribution. The samples in each of $X_1$, and $X_2$ are represented as $x_{s}^{1}$ and $x_{t}^{2}$ respectively. The number of samples in both $X_1$ and $X_2$ are equal. The samples in $Y$ are represented by $y_{s}$. Also, there exists a paired data correspondence between $X_1$ and $Y$, i.e. we have $\{(x_{s}^{1},y_{s})\}$. However, we have only $\{x_{t}^{2}\}$ and no corresponding data in $Y$. Finally, $X_1$ comes from a constrained setting, whereas $X_2$ has a great variety.

\noindent\textbf{Learning a Mapping: } We use the $\{(x_{s}^{1},y_{s})\}$ to learn a mapping $f$ for this data (Figure~\ref{fig:overview}-a). This mapping ($f$) is an example of paired image-to-image translation, and that we can minimize reconstruction error on paired data:
\begin{align}
  \min_{f} \sum_s ||y_s - f(x_{s}^{1})||_2
\label{eq:reg}
\end{align}

\noindent\textbf{Learning from Simulated Labels: } We do not have any paired data for $X_2$ (Figure~\ref{fig:overview}-b). We simulate the labels by using the samples $\{x_{t}^{2}\}$ and $f$ learned in Equation~\ref{eq:reg} (Figure~\ref{fig:overview}-c). This enables us to get a paired data between  $X_2$ and $Y$, $\{(x_{t}^{2}, f(x_{t}^{2})\}$. We intend to learn a \emph{new} mapping $g$ (Figure~\ref{fig:overview}-d) over the data pair $\{(x_{t}^{2}, f(x_{t}^{2})\}$. Since we now have labels for $X_2$, we can learn a mapping by minimizing reconstruction for the simulated data pair. 
\begin{align}
  \min_{g} \sum_t ||f(x_{t}^{2}) - g(x_{t}^{2})||_2
\label{eq:reg2}
\end{align}

\noindent More precisely, we are forcing $g$ to learn $f$ via samples in $X_2$, i.e. $\{x_{t}^{2}\}$. Importantly, the number of samples in $X_2$ are sufficient to learn the parameters of $f$.

\noindent\textbf{Our Observation: } We now have two mapping functions $f$ and $g$, where $g$ is trying to \emph{mimic} $f$ by learning over the samples of $X_2$. If there was no role of $X$ in learning this mapping, both $f$ and $g$ should behave similarly when fine-tuned for a particular task for different data sources. Infact, $g$ should underperform because it is an approximation of $f$. We make a test scenarios to see if this holds. We use $f$ and $g$ as an initialization for a task whose  data distribution is $X_1$ and learns a mapping to $Y$. Our findings suggest that $g$ perform better than $f$. This means that a representation learnt on a diverse data-distribution (even in the absence of paired data) act as a better initialization.

\subsection{Implementation Details}
\label{sec:implementation}

We now explain the different components that will be used in our experiments. We consider the task of surface normal estimation~\cite{Bansal16,Eigen15,Fouhey13a} for learning a representation as it naturally provides for the different data distribution described above. We use NYU-v2 depth dataset~\cite{Silberman12} and ImageNet~\cite{Russakovsky15} for our experiments. The NYU-v2 depth dataset~\cite{Silberman12} consists of $220,000$ video  frames collected using a Kinect in the indoor scenes. Each frame has a depth map that helps in computing a surface normal map. In our settings, the NYU-v2 depth dataset acts as source for $X_1$ and $Y$. We use a random subset of ImageNet~\cite{Russakovsky15} for $X_2$. This subset of ImageNet contains same number of images as in $X_1$. The ImageNet dataset provides a variety of images, and does not have any corresponding depth/surface-normal labeled data. The two data sources are quite complimentary as one is focussed primarily on the indoor scenes collected using a Kinect, the other is primarily a collection of web images that has a big proportion of outdoor scenes. Our goal in this work is to isolate the impact of \emph{visual data} and its diversity. To ensure this, we fix the number of images in two data sources as well to avoid any bias in our experiments.

\noindent\textbf{Default Model: } We use the model from Bansal \etal~\cite{PixelNet,Bansal16} for surface normal estimation. We briefly describe the model here. This network architecture, also known as PixelNet~\cite{PixelNet}, consists of a VGG-16 style architecture~\cite{SimonyanZ14a} and a multi-layer perceptron (MLP) on top of it for pixel-level prediction. There are $13$ convolutional layers and three fully connected (\emph{fc}) layers in VGG-16 architecture. The first two \emph{fcs} are transformed to convolutional filters following~\cite{Long15}. We denote these transformed {\em fc} layers of VGG-16 as conv-$6$ and conv-$7$. All the layers are denoted as  \{$1_1$, $1_2$, $2_1$, $2_2$, $3_1$, $3_2$, $3_3$, $4_1$, $4_2$, $4_3$, $5_1$, $5_2$, $5_3$, $6$, $7$\}. We use hypercolumn features from conv-\{$1_2$, $2_2$, $3_3$, $4_3$, $5_3$, $7$\}. An MLP is used over hypercolumn features with 3-fully connected layers of size $4,096$ followed by ReLU~\cite{krizhevsky2012imagenet} activations, where the last layer outputs predictions for $3$ outputs ($n_x$, $n_y$, $n_z$) with a euclidean loss for regression. Finally, we use batch normalization~\cite{Ioffe:2015} with each convolutional layer when training from scratch for faster convergence. More details about the architecture/model can be obtained from \cite{PixelNet}.

\noindent\textbf{Learning a mapping $f$: } We use the above model, initialize it with a random gaussian distribution, and train it for NYU-v2 depth dataset. The initial learning rate is set to $\epsilon = 0.001$, and it drops by a factor of 10 at step of $50,000$. The model is trained for $60,000$ iterations. We use all the parameters from ~\cite{PixelNet}, and have kept them fixed for all our experiments to avoid any bias due to hyper-parameter tuning.

\noindent\textbf{Learning a mapping $g$: } Firstly, we need to create labels for $X_2$ to learn $g$. We use $f$ trained above with the randomly subsampled images from ImageNet to create the training data pair. We use this data to learn mapping function $g$ that is trained from scratch and follows the same procedure as $f$.

\noindent\textbf{Using a million unlabelled images: } Finally, we use $f$ trained above with a million images from ImageNet to create the training data pair. We use this data to learn mapping function $h$ that is trained from scratch and follows the same procedure as $f$ (except that step size is now $200,000$ and we train it for $430,000$ iterations)\footnote{We arbitrarily shut the training of this model after 2 epochs. Better models may be learn by running it for longer.}.

We have tried to make sure that only thing that change in this experiment is the data source ($X_1$ and $X_2$), and rest everything is kept fixed to avoid any external influence on these experiments. We will now evaluate $f$, $g$, and $h$ for two tasks: (1). \textbf{Surface normal estimation} - We use mappings $f$, $g$, and $h$, and fine-tune them using NYU-v2 depth dataset~\cite{Silberman12} for surface normal estimation. We achieve better performing models as we condition the representation on diverse and more examples. Note that all the hyper-parameters and settings are kept same for analysis. (2). \textbf{Semantic Segmentation} - we use $h$ for semantic segmentation using the PASCAL VOC-2012 dataset~\cite{Everingham10} and achieve better results. We also improve the results further by going back to the unlabelled images and training a new representation.


\section{Analysis}
\label{sec:exp}

\begin{figure*}
\includegraphics[width=\linewidth]{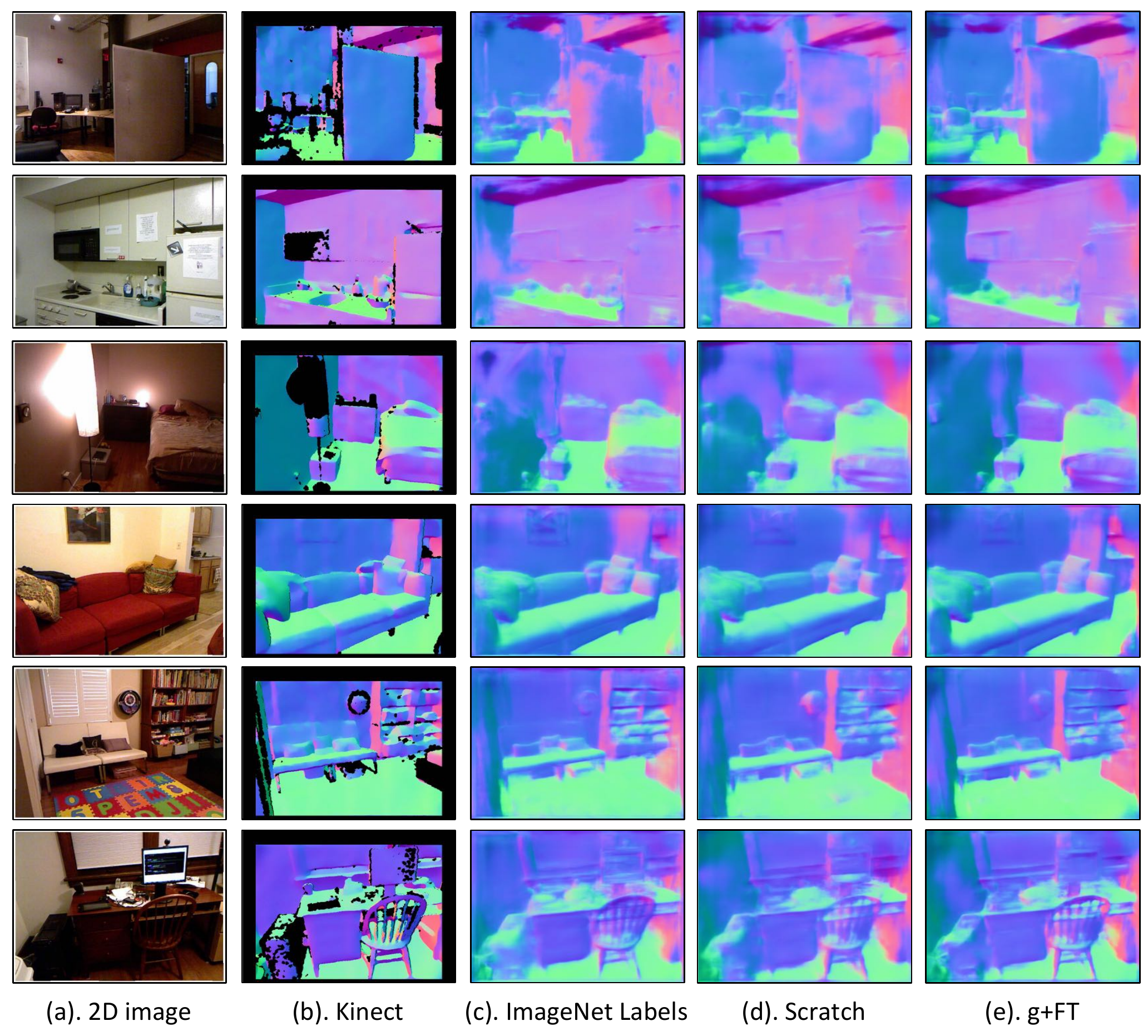}
\caption{\textbf{Influence of Unlabelled Data on Surface Normal Estimation: } For a given single 2D image (shown in \textbf{(a)}), we contrast the performance of various model. We  show the results from prior work~\cite{PixelNet,Bansal16} in \textbf{(c)}. This work use a pre-trained ImageNet classification model for initialization. We term it \textbf{ImageNet labels}. We show the outputs of the model trained from scratch \textbf{($f$)} in \textbf{(d)}. Finally we show our results \textbf{($g$+FT)} in \textbf{(e)}.  We can easily notice the influence of unlabelled data by contrasting results in \textbf{(d)} and \textbf{(e)}. This improvement in performance comes by conditioning the learnt representation on a diverse data and does not require any additional cost. For reference, we have also shown normals from \textbf{kinect} in \textbf{(b)}.}
\label{fig:normals}
\end{figure*}

We now quantitatively and qualitatively evaluate our hypothesis described in Section~\ref{sec:approach}. 

\subsection{Surface Normal Estimation}

We fine-tune $f$, $g$, and $h$ on NYU-v2 depth dataset~\cite{Silberman12} (described earlier in Section~\ref{sec:implementation}) for surface normal estimation. The initial learning rate is set to $\epsilon = 0.001$, and it drops by a factor of 10 at step of $50,000$. Each model is fine-tuned for $60,000$ iterations. We use 654 images from the test set of NYU-v2 depth dataset~\cite{Silberman12} for evaluation. Following \cite{Bansal16}, we compute six statistics over the angular error between the predicted normals and depth-based normals to evaluate the performance -- \textbf{Mean}, \textbf{Median}, \textbf{RMSE}, \textbf{11.25$^\circ$}, \textbf{22.5$^\circ$}, and \textbf{30$^\circ$} --  The first three criteria capture the mean, median, and RMSE of angular error, where lower is better. The last three criteria capture the percentage of pixels within a given angular error, where higher is better. 

Table~\ref{tab:data} compares the performance of $f$ and $g$ when fine-tuned on NYU-v2 for surface normal estimation. Each of them is denoted as $f$+FT, $g$+FT, and $h$+FT respectively. We observe that $f$+FT improves over $f$. More importantly, $g$+FT has a better performance than $f$+FT, and is comparable to the model fine-tuned from the ImageNet with class labels. Further, we observe that $f$+FT saturates and does not improve performance but $g$+FT when allowed to run for longer ($120,000$ iterations) could further improve the performance and can also get a performance better than a ImageNet (with class labels) pre-trained model. Further, with the increase in the number of unlabelled images ($h$ + FT), we can even achieve better performance. A recent work~\cite{qi2018geonet} gets similar performance by a careful use of multi-task optimization with a pre-trained ImageNet classification model. These results suggest that we can get a better performance with a small labeled data and millions of unlabelled images. More importantly, this experiment suggest that there is something peculiar with the visual data that enables us to get better performing models with the low performing models by just use of diverse unlabelled images. 

Figure~\ref{fig:normals} qualitatively compares the performance of different models. Our approach is able to correct the normals where the previous model failed, and could also get better outputs than prior art. 

\begin{table}
\scriptsize{
\setlength{\tabcolsep}{3pt}
\def\arraystretch{1.2}
\center
\begin{tabular}{@{}l  c c c c c c }
\toprule
\textbf{Approach}  &  Mean  &   Median & RMSE &  11.25$^\circ$ & 22.5$^\circ$ &  30$^\circ$ \\
\midrule
ImageNet Labels~\cite{Bansal16,PixelNet} &  19.8     &	12.0	 &   28.2	 &   47.9   &	   70.0  & 77.8	\\
\midrule
Scratch ($f$)  &   21.2 & 13.4   & 29.6 &  44.2 & 66.6  & 75.1  \\
\midrule
$f$ +FT        &   20.4 & 12.6   & 28.7 &  46.3 & 68.2  & 76.4  \\
$g$+FT        &  19.8 & 12.0   & 28.0 &  47.7 & 69.4  & 77.5  \\
$h$+FT        &  18.9 & 11.1   & 27.2 &  50.4 & 71.3  & 78.9  \\
\midrule
$g$+FT (until convergence)        &  19.4 & 11.5   & 27.8 &  49.2 & 70.4  & 78.1  \\
$h$+FT (until convergence)        &  \textbf{18.7} & \textbf{10.8}   & \textbf{27.2} &  \textbf{51.3} & \textbf{71.9}  & \textbf{79.3}  \\
\bottomrule
\end{tabular}
\vspace{0.2cm}
\caption{\textbf{Influence of Unlabelled Visual Data on Surface Normal Estimation: } We study the influence of data in this experiment. The top row shows the performance of surface normal estimation when a pre-trained ImageNet classification model is used for initialization. The second row shows the performance when a model is trained from scratch (initialized from a random gaussian distribution). This is the $f$ model in our setting. The next two rows shows $f$ and $g$ fine-tuned for NYU-v2 depth dataset for surface normal estimation. We observe that for same compute $g$+FT improves the performance over $f$+FT. Further, we observe that $f$+FT \emph{saturates} but $g$+FT improves and gets performance even better than prior work that use pre-trained ImageNet classification model (first row). Finally, we demonstrate as how performance can be further improved by using more unlabelled images. $h$+FT is trained using a million images in contrast to $g$+FT that is using $220,000$ images. We get 3-4\% better performance over base model.}
\label{tab:data}
}
\end{table}

\noindent\textbf{Do we improve globally or locally? } One may suspect that a model initialized with the weights of pre-trained ImageNet classification model may capture more local information as the pre-training consists of class labels.  We analyzed if $g$+FT is also able to capture these local aspects in the scene or is it capturing more global information. Table~\ref{tab:data2} contrast the performance of two approaches on indoor scene furniture categories such as chair, sofa, and bed. We observe that despite being trained on one-sixth of training data and without any explicit class labels, the performance of $g$+FT is competitive (and sometimes even slightly better) to the one using pre-trained ImageNet classification model). Finally, the performance for local objects exceeds prior art when trained using a million unlabelled images ($h$+FT). This suggests that we can capture both local and global information quite well without class-specific information.

\begin{table}
\scriptsize{
\setlength{\tabcolsep}{3pt}
\def\arraystretch{1.2}
\center
\begin{tabular}{@{}l  c c c c c c }
\toprule
\textbf{}  &  Mean  &   Median & RMSE &  11.25$^\circ$ & 22.5$^\circ$ &  30$^\circ$ \\
\midrule
\textbf{chair}  &   &   & &   &  &  \\
ImageNet Labels~\cite{Bansal16,PixelNet} &  31.7     &	24.0	 &   40.2	 &   \textbf{21.4}   &	   47.3  & 58.9	\\
$g$+FT (until convergence)      						 &  32.4     	     &      25.2  		 &  40.5 			&  19.1 	&   	45.2 &   57.3  \\
$h$+FT (until convergence)      						 &  \textbf{31.2}     	     &      \textbf{23.6}  		 &  \textbf{39.6} 			&  21.0	&   	\textbf{47.9} &   \textbf{59.8}  \\
\midrule
\textbf{sofa}  &   &   & &   &  &  \\
ImageNet Labels~\cite{Bansal16,PixelNet} &  20.6     &	15.7 &   26.7	 &   35.5  &	 66.8  & 78.2	\\
$g$+FT (until convergence)   						  &  21.4     &  16.1     &   27.6 &      34.9  &         64.4 &  76.1  \\
$h$+FT (until convergence) &  \textbf{20.0}     &	\textbf{15.2}	 &   \textbf{26.1}	 &   \textbf{37.5}  &	 \textbf{67.5}  & \textbf{79.4}	\\
\midrule
\textbf{bed}  &   &   & &   &  &  \\
ImageNet Labels~\cite{Bansal16,PixelNet} &  19.3     &	13.1	 &   26.6	 &  44.0   &	   70.2 &  80.0	\\
$g$+FT (until convergence)       	&  19.2 		   & 12.9   			& 26.4 			& 44.6		 & 	70.3		  & 79.7 \\
$h$+FT (until convergence)       	&  \textbf{18.4} 		   & \textbf{12.3}   			& \textbf{25.5} 			& \textbf{46.5}		 & 	\textbf{72.7}		  & \textbf{81.7} \\
\bottomrule
\end{tabular}
\vspace{0.2cm}
\caption{\textbf{Performance for local objects: } We contrast the performance of our approach with the model fine-tuned using ImageNet (with class labels) on furniture categories, i.e. chair, sofa, and bed. We observe that $g$ is  competitive (and sometimes even slightly better) to prior art even though it did not use any explicit class information that is available to a pre-trained ImageNet classification model. Finally, our approach exceeds the performance of prior art when using a million images ($h$+FT).}
\label{tab:data2}
}
\end{table}

\begin{table*}[t!]
\scriptsize{
\setlength{\tabcolsep}{3pt}
\def\arraystretch{1.2}
\center
\begin{tabular}{@{}l c c c c c c c c c c c c c c c c c c c c c c}
\toprule
\textbf{VOC 2012 test}  & aero  &   bike &  bird & boat &  bottle  &  bus  &  car  &  cat  &  chair & cow & table  &  dog  & horse & mbike & person  & plant & sheep & sofa & train & tv &  bg &\textbf{mAP} \\
\midrule
Scratch~\cite{PixelNet}  & 62.3  &   26.8 &  41.4 & 34.9 &  44.8  &  72.2  &  59.5  &  56.0  &  16.2 & 49.9 & 45.0 & 49.7  & 53.3 & 63.6 & 65.4   & 26.5 & 46.9 & 37.6 & 57.0 & 40.4 & 85.2  &\textbf{49.3}\\
\midrule
Geometry~\cite{PixelNet}  & 71.8  &  29.7 & 51.8 & 42.1 &  47.8  &  77.9  &  65.9  &  59.7  &  19.7 & 50.8 & 45.9 &  55.0  & 59.1 & 68.2 & 69.3  & 32.5 & 54.3 & 42.1 & 60.8 &43.8 & 87.6 &\textbf{54.1} \\
Our Approach ($h$)  & 74.4  &   34.5 &  60.5 & 47.3 &  57.1  &  74.3  &  73.1  &  61.7  &  22.4 & 51.4 & 36.4 &  52.0  & 60.9 & 68.5 & 69.1  & 37.6 & 58.0 & 34.3 & 64.3 & 50.2 & 90.0  &\textbf{56.1}\\
+ Final   & 82.2 & 35.1  & 62.0 & 47.4 & 62.1   & 76.6  & 74.1   & 62.7  & 23.9  & 49.9 & 47.0 & 55.5   & 58.0  & 74.9 & 73.9  & 40.1 & 56.4 & 43.6  & 65.4  & 52.8 & 90.9  & \textbf{58.8} \\
\midrule
ImageNet~\cite{PixelNet}  & 79.0  &   33.5 &  69.4 & 51.7 &  66.8  &  79.3  &  75.8  &  72.4  &  25.1 & 57.8 & 52.0 &  65.8  & 68.2 & 71.2 & 74.0  & 44.1 & 63.7 & 43.4 & 69.3 & 56.4 & 91.1  &\underline{\textbf{62.4}}\\
\bottomrule
\end{tabular}
\newline
\vspace{0.2cm}
\caption{\textbf{Evaluation on VOC-2012: } We compare the performance of model fine-tuned from $h$ with the model trained from scratch (random gaussian initialization). We observe a significant 7\% improvement in performance. We also compare with the prior work~\cite{PixelNet} that used models trained from normals for NYU-v2 as an initialization. We observe a 2\% improvement in performance just by changing the underlying data to learn the representation. We further improve the performance by $2.7$\% by running the previous model on unlabelled images, and training a model from scratch specifically for segmentation. Finally, we observe that our approach has closed the gap between ImageNet (with class labels) pre-trained model and self-supervised model to 3.6\%.}
\label{tab:voc_2012}
}
\end{table*}

\subsection{Semantic Segmentation}

We now evaluate $h$ for the task of semantic segmentation. We fine-tune $h$ using the training images from PASCAL VOC-2012~\cite{Everingham10} for semantic segmentation, and additional labels collected on 8498 images by Hariharan et al.~\cite{hariharan11}. We evaluate the performance on the test set that required submission on PASCAL web server~\cite{pascal}. We report results using the standard metrics of region intersection over union (\textbf{IoU}) averaged over classes (higher is better). 

We follow \cite{PixelNet} for this experiment. The initial learning rate is set to $\epsilon = 0.001$, and it drops by a factor of 10 at step of $100,000$. The model is fine-tuned for $160,000$ iterations.  Table~\ref{tab:voc_2012} contrasts the performance of our approach with other approaches. We observe a slight performance improvement over the prior work~\cite{PixelNet} that used normals for initialization, and 7\% over the model trained from scratch. Finally, we follow the approach similar to surface normal estimation. We ran the trained model on a million unlabelled images, train a new model from scratch for segmentation\footnote{We used a batch-size of 5. The initial learning rate is set to $\epsilon = 0.001$, and it drops by a factor of 10 at step of $250,000$. The model is trained for $300,000$ iterations. More iterations may further help in improving performance. Finally, there may be a better choice of hyper-parameters that can give more boost in performance. We have not explored that space.}, and fine-tune it for PASCAL dataset (using same hyper-parameters as earlier). We observe a further $2.7$\% boost in the performance thereby closing the gap between a pre-trained model and self-supervised model to 3.6\%. We hope that use of more unlabelled images (probably ten or hundred millions) can drastically improve the performance.

\section{Discussion}
\label{sec:discussion}

The current experiments suggest that using a million unlabelled images from ImageNet can help us get better performing models.  Our observations are currently limited to surface normal estimation and semantic segmentation. Our choice of the task was primarily motivated by the two different data distribution available for this task.  However, we hope that our work inspires the community to conduct experiments for more tasks especially the pixel-level tasks, where it is hard to collect the ground truth data. There is no scarcity of images available on web, and that it seems we can improve the performance without any additional expense of labeling. 

\vspace{.5cm}
{\noindent\textbf{Acknowledgements: }  I would like to thank Xiaolong Wang for the motivation to upload this technical report to arXiv.

%

\end{document}